\pdfoutput=1
\documentclass{bmvc2k}
\usepackage[latin9]{inputenc}
\usepackage{geometry}
\usepackage{color}
\usepackage{verbatim}
\usepackage{amsmath}
\usepackage{amssymb}

\makeatletter

\usepackage{graphicx}
\usepackage{amsmath,amssymb} 

\usepackage{color}
\usepackage{array}
\usepackage{multirow}
\usepackage{amsmath}
\usepackage{graphicx}
\usepackage{color}

\makeatother

\title{Transductive Multi-label Zero-shot Learning}

\addauthor{Yanwei Fu, Yongxin Yang}{{y.fu, yongxin.yang}@qmul.ac.uk}{0}
\addauthor{Timothy Hospedales, Tao Xiang}{{t.hospedales, t.xiang}@qmul.ac.uk}{0}
\addauthor{Shaogang Gong}{{s.gong}@qmul.ac.uk}{0}

\addinstitution{
 School of EECS\\
 Queen Mary University of London\\
 London, E1 4NS, UK
}

\runninghead{Fu ET AL}{Transductive Multi-label Zero-shot Learning}

\begin{document}



\maketitle
\begin{abstract}
Zero-shot learning has received increasing interest as a means to
alleviate the often prohibitive expense of annotating training data
for large scale recognition problems. These methods have achieved
great success via learning intermediate semantic representations in
the form of attributes and more recently, semantic word vectors. However,
they have thus far been constrained to the single-label case, in contrast
to the growing popularity and importance of more realistic multi-label
data. In this paper, for the first time,  we investigate and formalise a general framework for multi-label zero-shot
learning, addressing the unique challenge therein:
how to exploit multi-label correlation at test time with no training
data for those classes? In particular,  we propose (1) a multi-output deep regression model to project an image into a semantic word space, which explicitly exploits the correlations
in the intermediate semantic layer of word vectors; (2) a novel
zero-shot learning algorithm for multi-label data that exploits the unique compositionality
property of semantic word vector representations; and (3) a transductive learning
strategy to enable the regression model learned from seen classes to generalise well to unseen classes. Our zero-shot learning experiments on a number of  standard multi-label
datasets demonstrate that our method outperforms a variety of baselines.
\end{abstract}

\section{Introduction}

There are around 30,000 human-distinguishable basic object classes
\cite{object_cat_1987} and many more subordinate ones. %
 A major barrier to progress in visual recognition is thus collecting
training data for many classes. Zero-shot learning (ZSL) strategies have
therefore gained increasing interest as a route to side-step this
prohibitive cost, as well as enabling potential new categories emerging
over time to be represented and recognised. To classify instances
from a class with no examples, ZSL exploits knowledge transferred from
a set of seen (auxiliary) classes to unseen (test) classes, typically
via an intermediate semantic representation such as attributes. This
has recently been explored at large scale
on ImageNet \cite{DeviseNIPS13,RohrbachCVPR12}.

Prior zero-shot learning
methods have assumed that class labels on each instance are mutually
exclusive, i.e., multi-class single label classification. Nevertheless many real-world data
are intrinsically multi-label. For example, an image on Flickr often contains multiple objects with cluttered background, thus requiring more than one label to describe its content. There is an even more acute need for zero-shot learning in the case of multi-label classification. This is because different labels are often correlated (e.g.~cows often appear on grass). In order to better predict these labels given an image, the label correlation must be modelled. However,  for $n$ labels, there are $2^n$ possible multi-label combinations and to collect sufficient training samples for each combination to learn the correlations of labels is infeasible. 
It is thus surprising to note that there is little if any existing work on multi-label zero-shot learning. Is it because there is a trivial extension of existing single label ZSL approaches to this new problem? By assuming each label is independent from one another, it is indeed possible to decompose a multi-label ZSL problem into multiple single label ZSL problems and solve them using existing single label ZSL methods. However this does not exploit label correlation, and we demonstrate in this work that this naive extension leads to very poor label prediction for unseen classes. Any attempt to model this correlation, in particular for the unseen classes with zero-shot, is extremely challenging.

In this paper, a novel framework for multi-label zero-shot learning is proposed. Our framework is based on transfer learning -- given a training/auxiliary dataset containing labelled images, and a test/target dataset with a set of unseen labels/classes (i.e.~none of the labels appear in the training set), we aim to learn a multi-label classification model from the training set and generalise/transfer it to the test set with unseen labels. This knowledge transfer is achieved using an intermediate semantic representation in the form of the skip-gram word vectors  \cite{wordvectorICLR,distributedword2vec2013NIPS} learned from linguistic knowledge bases. This representation is shared between the training and test classes, thus making the transfer possible.

More specifically, our framework has two main
components: multi-output deep regression (Mul-DR) and  zero-shot multi-label
prediction (ZS-MLP). Mul-DR is a 9 layer neural
network that exploits the widely used convolutional neural network
(CNN) layers \cite{RazavianICLR2014}, and includes two multi-output regression layers as the final layers. It learns
from auxiliary data the explicit and direct mapping from raw image pixels to a linguistic representation
defined by the skip-gram language model \cite{wordvectorICLR,distributedword2vec2013NIPS}. With Mul-DR, each test image is now projected into the semantic word space where the unseen labels and their combinations can be represented as data points without the need to collect any visual data.
ZS-MLP aims to address the multi-label ZSL problem in this semantic word space. Specifically, we note that in this space any label combination can be synthesised.
We thus exhaustively synthesise the power set of all possible prototypes
(i.e., combinations of multi-labels) to be treated as if they were a  set of labelled
instances in the space. With this synthetic dataset, we are able to
extend conventional multi-label algorithms \cite{TransdMultilabel,10.1109/TKDE.2013.39,wu:multi-label, Multilabel2012ZSL},
to propose two new multi-label algorithms -- direct multi-label zero-shot
 prediction (DMP) and transductive multi-label zero-shot prediction
(TraMP). However, since Mul-DR is learned using the auxiliary classes/labels, it may not generalise well to the unseen classes/labels. To overcome this problem, we further exploit self-training to adapt the Mul-DR to the test classes to improve its generalisation capability.

\section{Related Work}

\textbf{Multi-label classification}\quad Multi-label classification
has been widely studied -- for a review of the field please see \cite{10.1109/TKDE.2013.39,wu:multi-label}. Most previous
studies assume plenty of training data. Recently efforts have been made to relax this assumption. Kong
\emph{et al.} \cite{TransdMultilabel} studied transductive multi-label
learning with a small set of  training instances.  Hariharan \emph{et al.}
\cite{Multilabel2012ZSL} explored the label correlations of auxiliary
data via a multi-label max-margin formulation and better incorporated
such label correlations as prior for multi-class zero-shot learning
problem. However, none of them addresses the  multi-label zero-shot
learning problem tackled in this work.

\noindent\textbf{Zero-shot learning}\quad Multi-class single label zero-shot learning
has now been widely studied using attribute-based intermediate
semantic layers \cite{ferrari2007attrib_learn,palatucci2009zero_shot,lampert2009zeroshot_dat,
yanweiembedding, yanweiranking, crowdcountingKE} or data-driven \cite{yanweiPAMIlatentattrib,yanweiUSAA,lampert2012augmentedAttribs,layne2014wildAttr}
representations. However attribute-based strategies have limited ability to scale
to many classes because the attribute ontology has to be manually defined. To address this limitation,
Socher \emph{et al.} \cite{RichardNIPS13} first employed a linguistic
model \cite{UnsupervisedLangModel2012} as the intermediate semantic
representation. However, this does not model the syntactic
and semantic regularities in language \cite{distributedword2vec2013NIPS}
which allows
vector-oriented reasoning. Such a reasoning is critical for our ZS-MLP to synthesise label combination prototypes in the semantic word space. For example,
$Vec(``Moscow")$ should be much closer to $Vec(``Russia")+Vec(``capital")$ than
$Vec(``Russia")$ or $Vec(``capital")$ only.   For this purpose, we employ
 the skip-gram language model to learn the word space, which has shown to be able to capture such syntactic regularities \cite{wordvectorICLR,distributedword2vec2013NIPS}. Frome \emph{et al.}  \cite{DeviseNIPS13} also used the  skip-gram language model. They learned a visual-semantic embedding model --
DeViSE model for single label zero-shot learning by projecting both
visual and semantic information of auxiliary data into a common space. However there are a number of fundamental differences between their work and ours: (1) Comparing the DeViSE model with our Mul-DR, the learning of the mapping between images and the semantic word space by Mul-DR is more explicit and direct. We show in our experiments that this leads to better projections and thus better classification performance. (2) Our Mul-DR can generalise better to the unseen test classes thanks to our self-training based transductive learning strategy. (3) Most critically, we address the multi-label ZSL problem whilst they only focused on the single label ZSL problem. Additionally, zero-shot learning can be taken as the generalisation of class-incremental learning (C-IL)~\cite{Zhou2002515, CIL_augumented} or life-long learning~\cite{lifelonglearning}.

\noindent\textbf{Our Contributions}\quad Overall, we make following
contributions: (1) As far as we know this is the first work that addresses the  multi-label
zero-shot learning problem. (2) Our multi-output deep regression framework exploits
correlations across dimensions while learning the direct mapping from images
to intermediate skip-gram linguistic word space. (3) Within the linguistic
space, two algorithms are proposed for multi-label ZSL. (4) We propose
a simple self-training strategy to make the deep regression model generalise better to the unseen test classes.  (5) Experimental
results on benchmark multi-label datasets show the efficacy of our framework for multi-label ZSL over a variety of baselines.

\vspace{-0.4cm}

\section{Methodology}

\subsection{Problem setup}

Suppose we have two datasets -- source/auxiliary and target/test.
The auxiliary dataset $S=\left\{ X_{S},Y_{S},L_{S},\mathcal{W}_{S}\right\} $
has $n_{S}$ training instances and test dataset $T=\left\{ X_{T},Y_{T},L_{T},\mathcal{W}_{T}\right\} $
has $n_{T}$ test instances. We use $\mathcal{S}=\left\{ 1,\cdots,n_{S}\right\} $ and $\mathcal{U}=\left\{ n_{S}+1,\cdots,n_{T}+n_{S}\right\} $
to denote the index set for instances in auxiliary and test dataset.  $X_{S}=\left\{ \mathbf{x}_{1},\cdots,\mathbf{x}_{n_{S}}\right\} $
and $X_{T}=\left\{ \mathbf{x}_{n_{S}+1},\cdots,\mathbf{x}_{n_{S}+n_{T}}\right\} $
are the raw image data of all auxiliary and test instances respectively.
 $Y_{S}=\left[\mathbf{y}_{1},\cdots,\mathbf{y}_{n_{S}}\right]$ and $Y_{T}=\left[\mathbf{y}_{n_{S}+1},\cdots,\mathbf{y}_{n_{S}+n_{T}}\right]$
are the intermediate semantic representations of each auxiliary and
test instance -- in our case $\mathbf{y}_{i}$ is a $100$ dimensional continuous word
vector for instance $i$ in the skip-gram language model \cite{distributedword2vec2013NIPS}
space. $L_{s}=\left[\mathbf{l}_{1},\cdots,\mathbf{l}_{n_{S}}\right]$ and $L_{T}=\left[\mathbf{l}_{n_{S}+1},\cdots,\mathbf{l}_{n_{S}+n_{T}}\right]$ are the label vectors for auxiliary and test dataset to be predicted respectively.

The possible \emph{textual} labels for each instance
in $L_{S}$ and $L_{T}$ are denoted $\mathcal{W}_{S}=\left\{ w_{1},\cdots,w_{m_{S}}\right\} $ and $\mathcal{W}_{T}=\left\{ w_{m_{S}+1},\cdots,w_{m_{S}+m_{T}}\right\} $ respectively,
where $m_{S}$ and $m_{T}$ are the total number of classes/labels in each
dataset.  Given a label-space of $m_{T}$ binary labels, an instance $\mathbf{x}_{i}$
can be tagged with any of the $2^{m_{T}}$ possible label subsets,
$\mathbf{l}_{i}\in\{0,1\}^{2^{m_{T}}}$, where $\mathbf{l}_{ij}=1$
means instance $i$ has label $j$, and $\mathbf{l}_{ij}=0$ means
otherwise. Denoting the power sets of textual labels $\mathcal{W}_{S}$
and $\mathcal{W}_{T}$ as $\mathcal{P}\left(\mathcal{W}_{S}\right)$
and $\mathcal{P}\left(\mathcal{W}_{T}\right)$, for multi-label
classification we need to find the optimal class label set
column vector $\mathbf{l}_{i}$ for the $i-th$ test instance in
the power set space $\mathcal{P}\left(\mathcal{W}_{T}\right)$. At
training time $X_{S},Y_{S},L_{S},\mathcal{W}_{S}$ are all observed.
At test time only new class names $\mathcal{W}_{T}$ and images
 $X_{T}$ are given, their representation $Y_{T}$ and multi-label
vectors $L_{T}$ are to be predicted.

\subsection{Learning a semantic word space }
\label{sub:Learning-Semantic-Word}

The semantic representations $Y_{S}$ and $Y_{T}$ are the projection
of each instance into a linguistic word vector space $\mathcal{V}$.
The semantic word vector space is learned by using the state-of-the-art skip-gram
language model \cite{wordvectorICLR,distributedword2vec2013NIPS}
on all English Wikipedia articles%
\footnote{Only articles are used without any user talk/discussion. To 13 Feb.
2014, it includes 2.9 billion words and 4.33 million vocabulary (single
and bi/tri-gram words).%
}. The space $\mathcal{V}$ represents almost all available English
vocabulary and thus is potentially much more effective than human annotators to
measure subtle similarities and differences between any two textual
labels. Furthermore, $\mathcal{V}$ encodes  the syntactic
and semantic regularities in language \cite{distributedword2vec2013NIPS}
which allows
vector-oriented reasoning by its `compositionality' property. This property enables the
critical capability of synthesising the exhaustive set of
 test label combinations  $\mathcal{P}\left(\mathcal{W}_{T}\right)$.    Note that cosine distance
is used in the space $\mathcal{V}$ because of its robustness against
noise \cite{wordvectorICLR,distributedword2vec2013NIPS}. We use $v:\mathcal{W}\rightarrow\mathcal{V}$
to represent the skip-gram projection from textual concepts (words) in $\mathcal{W}$
to vectors in $\mathcal{V}$. Such a semantic space thus captures the correlations between labels without any need to collect visual examples --
the meaning of multiple labels for one instance can be inferred
by the sum of the word vector projections of its individual labels.
Formally, we have
\begin{align}
Y_{S} & =v(\mathcal{W}_{S})\cdot L_{S},~~~~~~Y_{T}\:=v(\mathcal{W}_{T})\cdot L_{T}\label{eq:label_S}
\end{align}
where $v\left(\mathcal{W}_{S}\right)$ and $v\left(\mathcal{W}_{T}\right)$
are the word vector projections of the label class sets in the auxiliary
and test datasets respectively. The next section discusses how to learn a predictive
model for $Y_{T}$ given visual data $X_{T}$.

\subsection{Multi-output deep regression}

We design a multi-output deep regression (Mul-DR) model $f:\mathcal{X}\to\mathcal{V}$
to predict the semantic representation $Y_{T}\in\mathcal{V}$ from images
$X_{T}\in\mathcal{X}$ where $\mathcal{X}$ is the space of raw image
pixel intensity values. Our Mul-DR is inspired by the recent success of  the deep convolutional
neural network (CNN) features \cite{imagenetdeeplearning,overfeat} as well as the importance of modelling
correlations within the semantic representation.
The Mul-DR model is a neural network composed of nine layers: Layer $1-5$
are convolutional layers; Layer $6-8$ are fully connected layers;
Layer $9$ is the linear mapping layer with $100$ least square regressors.

Two key components contribute to the effectiveness of Mul-DR. The first component (layers 1-7) provides state-of-the-art feature
extraction for many computer vision tasks
\cite{RazavianICLR2014}. It directly maps the raw image to the powerful CNN features\footnote{However, it has more than 148.3 millions parameters and thus to prevent
overfitting on small auxiliary dataset, ImageNet with 1.2 million
labelled instances are used to train this component~\cite{overfeat}.%
}, avoiding the pitful of bad performance due to ``wrong selection''
of features for a given dataset. The second component (layers 8-9) provides the multi-output neural network
(NN) regressors. Different from \cite{imagenetdeeplearning,overfeat},
where the 8-th layer is an output layer for classification,  the 8-th layer in our model is a fully
connected layer of 1024 neurons with Rectified Linear Units (ReLUs) activation functions. This
soft-thresholding non-linearity has better properties for generalisation
than the widely used tanh activation units.
Such a fully connected layer helps explore correlations among the different
dimensions in the semantic word space. The final (9-th) layer  of least square regressors provide an estimation of the 100 dimensional semantic representation in the space
$\mathcal{V}$.

 To apply this neural network, we resize all images $X_{S}$ and
$X_{T}$ to $231\times231$ pixels. The parameters of the first components
are pre-trained using ImageNet \cite{overfeat} while the parameters
of the second component are trained by gradient descendent with auxiliary
data $X_{S}$ and $Y_{S}$. At test time, Mul-DR predicts the semantic
word vector $\hat{\mathbf{y}_{i}}$ %
for each unseen image $\mathbf{x}_{i}\in X_{T},i\in\mathcal{U}$. Here the hat operator indicates the variable is estimated.

\subsection{Zero-shot multi-label prediction}

Given the estimated semantic representation $\hat{Y}_{T}$, we need
to infer the labels $\hat{L}_{T}$ of the test set. A straightforward
solution is to decompose the multi-label classification problem into multiple independent
binary classification problems which is equivalent \cite{hastie01statisticallearning}
to directly solving Eq (\ref{eq:label_S}) by:
\begin{equation}
\hat{L}_{T}=\left[\left[v(\mathcal{W}_{T})\right]^{T}v(\mathcal{W}_{T})\right]^{\dagger}\left[v(\mathcal{W}_{T})\right]^{T}\cdot\hat{Y}_{T}\label{eq:extendDAP1}
\end{equation}
where $\dagger$ is the Moore-Penrose pseudo-inverse. Eq (\ref{eq:extendDAP1})
directly predicts the labels of each instance by a linear transformation
of the intermediate representation $\hat{Y}_{T}$.
In a way, this can be considered as an extension of the `Direct Attribute Prediction (DAP)' \cite{lampert2009zeroshot_dat}
 to the case of multi-label and continuous representation. We thus term this method exDAP.
However, this does not exploit the multi-label correlations and thus has very limited expressive
power \cite{Zhang2007,rankSVMMultilabel}.
Hence we propose two more principled multi-label zero-shot algorithms -- Direct Multi-label zero-shot Prediction (DMP) and Transductive Multi-label zero-shot
Prediction(TraMP).

\noindent \textbf{Direct Multi-label zero-shot Prediction (DMP)}\quad Thanks
to the compositionality property  of $\mathcal{V}$, label-correlation can be
explored by synthesising the representation of every possible multi-label
annotations in $\mathcal{V}$: that is the power set of label vector
matrix $P=v\left(\mathcal{P}\left(\mathcal{W}_{T}\right)\right)$
where $P=\left[\mathbf{p}_{1},\cdots,\mathbf{p}_{2^{m_{T}}}\right]$.
Thus Eq (\ref{eq:extendDAP1}) is  replaced by a nearest neighbour (NN)
classifier using all the synthesised instances as training data. The
label set $\mathbf{l}_{i}$ of instance $i\in\mathcal{U}$ with representation
$\hat{\mathbf{y}}_{i}=f(\mathbf{x}_{i})$ is then assigned as $\mathbf{p}_{a}\in v\left(\mathcal{P}(\mathcal{W}_{T})\right)$,
where $a$ is the index computed by
\begin{equation}
a=\underset{j}{\mathrm{argmin}}\parallel\hat{\mathbf{y}}_{i}-\mathbf{p}_{j}\parallel\label{eq:NNclassifier}
\end{equation}
\noindent where $\parallel\cdot\parallel$ refers to the cosine distance.

\noindent \textbf{Transductive Multi-label zero-shot Prediction (TraMP)}\quad DMP
can explore label correlations but only insofar as encoded by the compositionality
of the prototypes in $\mathcal{V}$. It would be more desirable if the
manifold structure of $\hat{Y}_{T}$ given test instances $X_{T}$
could be used to improve multi-label zero-shot learning, i.e.~via transductive learning. We therefore
propose TramMP, which can be viewed as an extension the TRAM model in \cite{TransdMultilabel}
for zero-shot learning, or a semi-supervised generalisation of Eq (\ref{eq:NNclassifier}).
The key idea is to use the power set of prototypes $P$ as a known label
set and to perform transductive label propagation from $P$ to the inferred
semantic representations $\hat{Y}_{T}$. We denote the index of the
power set prototypes as $\mathcal{L}=\left\{ n_{S}+n_{T}+1,\cdots,n_{S}+n_{T}+2^{m_{T}}\right\} $
and its corresponding class label set as $L_{P}$. Specifically,
we define a k-nearest neighbour (kNN) graph among the  test instances
$\hat{Y}_{T}$ and prototypes $P$. For any two instances $i$ and
$z$, where $i,z\in\left\{ \mathcal{U},\mathcal{L}\right\} $,
\begin{equation}
\omega_{iz}=\begin{cases}
\begin{array}[t]{cc}
\frac{1}{Z_{i}}\mathrm{exp}\left(-\frac{\parallel\hat{\mathbf{y}}_{i}-\hat{\mathbf{y}}_{z}\parallel^{2}}{2\sigma^{2}}\right), & if\:{\color{black}{\color{blue}{\color{black}z\in NN_{k}\left(\hat{\mathbf{y}}_{i},\left[\hat{Y}_{T},P\right]\right)}}}\\
0 & otherwise
\end{array}\end{cases}\label{eq:kNNweight}
\end{equation}
\noindent where $\sigma\thickapprox\underset{i,z=1,\cdots,\mid\left\{ \mathcal{U},\mathcal{L}\right\} \mid}{\mathrm{median}}\parallel\hat{\mathbf{y}}_{i}-\hat{\mathbf{y}}_{z}\parallel^{2}$.
$NN_{k}\left(\hat{\mathbf{y}_{i}},\left[\hat{Y}_{T},P\right]\right)$ indicates
the index set of k-nearest neighbors of $\hat{\mathbf{y}}_{i}$ from $\left[\hat{Y}_{T},P\right]$.
$Z_{i}=\sum_{z\in NN_{k}\left(\hat{\mathbf{y}_{i}},\left[\hat{Y}_{T},P\right]\right)}\mathrm{exp}\left(-\frac{\parallel\hat{\mathbf{y}}_{i}-\hat{\mathbf{y}}_{z}\parallel^{2}}{2\sigma^{2}}\right)$
is the normalisation term to make sure $\sum_{z}\omega_{iz}=1$. We
define $A=I-\omega$ and partition the matrix $A$ into blocks,
$A=\left[\begin{array}{cc}
A_{\mathcal{L}\mathcal{L}} & A_{\mathcal{L}\mathcal{U}}\\
A_{\mathcal{U}\mathcal{L}} & A_{\mathcal{U}\mathcal{U}}
\end{array}\right]$
and the label set of test instances can be inferred by the following
closed form solution \cite{TransdMultilabel},
\begin{equation}
\hat{L}_{T}=-A_{\mathcal{U}\mathcal{U}}^{-1}A_{\mathcal{U}\mathcal{L}}L_{P}.
\label{eq:transductiveLP}
\end{equation}

\subsection{Generalisation of multi-output deep regression}

As described above, our framework consists of two key steps: applying the multi-output deep regression (Mul-DR) model to obtain the estimated semantic representation $\hat{Y}_{T}$, and followed by applying either DMP or TraMP to predict $L_{T}$. There is however an unsolved issue, that is, our Mul-DR is learned from the auxiliary data with a different set of labels from the target/test data. This projection model is thus not guaranteed to accurately project a test image to be near its ground truth label vector in the semantic word space. For example, if our Mul-DR is learned to project images of cat and dog to the word vector representation of ``cat" and ``dog" ($v(``cat")$ and $v(``dog")$), it may not accurately project an image with a person and a chair to its word vector representation of $v(``person")+v(``chair")$ when both labels were not available for learning the Mul-DR model. Any regression model will have such a generalisation problem especially when the test data are  distributed differently from the auxiliary data. To make the Mul-DR model generalise better to the target domain, we transductively exploit the predicted semantic representation
$\hat{Y}_{T}$ to update the power set of label vector matrix $P$. In this way the target data would be better aligned with the synthesised label combination vectors in the semantic word space, thus helping generalise the Mul-DR to the target domain.
This can be viewed as a semi-supervised learning (SSL) method starting
from one instance for each label combination if the synthesised prototypes themselves
are treated as instances. We therefore take a simple SSL strategy
and perform one step of self-training \cite{yanweiPAMIlatentattrib}
to refine each prototype of $P$,
\begin{eqnarray}
\overline{p}_{i} & = & \frac{1}{k}\sum_{\hat{\mathbf{y}}_{T}\in NN_{k}(p_{i},\hat{Y}_{T})}\hat{\mathbf{y}}_{T}\label{eq:self-training}
\end{eqnarray}
where $\bar{P}=\left[\bar{p}_{1},\cdots,\bar{p}_{2^{m_{T}}}\right]$
is the updated prototype matrix and $k$ is the number of nearest neighbour\footnote{Note that $k$ is not necessarily with the same $k$ value in Eq (\ref{eq:kNNweight}).}
selected. We use the updated label vector matrix $\bar{P}$ to compute
DMP (Eq (\ref{eq:NNclassifier})) and TramMP (Eqs (\ref{eq:kNNweight}) and (\ref{eq:transductiveLP}))
in our framework.

\section{Experiments}
\vspace{-0.2cm}

\textbf{Datasets}\quad Two popular multi-label datasets -- Natural
Scene \cite{Zhang2007} and IAPRTC-12 \cite{IAPRTC12} are used to
evaluate our framework. \textbf{Natural Scene} consists
of $2000$ natural scene images where each image can be labelled as
any combinations of \emph{desert}, \emph{mountains}, \emph{sea}, \emph{sunset}
and \emph{trees} and over $22\%$ of the whole dataset is multi-labelled.
For multi-label zero-shot learning on Natural Scene, we use a multi-class single label
dataset -- Scene dataset \cite{scene_OSR} (totally $2688$ images)
as the auxiliary dataset which have been labelled with a non-overlapping set of labels such as \emph{street}, \emph{coast} and \emph{highway}.  \textbf{IAPRTC-12} consists of $20000$ images
and a total of $275$ different labels. The labels are hierarchically
organised into $6$ main branches: humans, animals, food, landscape-nature,
man-made and other. Our experiments consider the subset of
landscape-nature branch (around $9500$ images) and use the top $8$ most frequent labels from this branch with over $30\%$ of multi-label test images. For zero-shot classification on this dataset, we employ both Scene and Natural Scene as the auxiliary dataset.

\vspace{-0.4cm}

\subsection{Experimental setup}

\noindent \textbf{Evaluation metrics}\quad
(a) \textbf{Hamming Loss}: it measures the percentage of mismatches
between estimated and ground-truth labels; (b) \textbf{MicroF1} \cite{correlatedML2006}:
it evaluates both micro average of Precision (Micro-Precision) and micro
average of Recall (Micro-Recall) with equal importance; (c) \textbf{Ranking
Loss}: given the ranked list of predicted labels, it measures the
number of label pairs that are incorrectly ordered by comparing their
confidence scores with the ground-truth labels; (d) \textbf{Average
precision}: given a ranked list of classes, it measures the area under
precision-recall curve. These four criteria evaluate very different aspects
of  multi-label classification performance. Usually very few algorithms
can achieve the best performance on all metrics. High values are preferred for MicroF1
and AP and vice-versa for Ranking and Hamming loss. For ease of interpretation
we present $1-$MicroF1 and $1-$AP; so smaller values for all
metrics are preferred.

\noindent \textbf{Competitors}\quad Our full framework
includes two main novel components: Mul-DR and DMP/TraMP. To evaluate
the effectiveness of these two components, we define several competitors
by replacing each component with possible alternatives. (1) \textbf{SVR+exDAP}:
Support Vector Regression (SVR)\footnote{For fair comparison, we use the CNN features output by the first component
(Layer 1-7) of our Mul-DR framework as the low-level feature for linear SVR used with the cost parameter set to 10.}~\cite{chang2001libsvm} is used to learn $f:\mathcal{X}\to\mathcal{V}$
and infer the representation of each test instance. Using exDAP (Eq (\ref{eq:extendDAP1})) is a straightforward generalisation of \cite{lampert2009zeroshot_dat,lampert13AwAPAMI} to multi-label zero-shot learning.  (2) \textbf{SVR+DMP}: SVR replaces
Mul-DR and we further use DMP (Eq (\ref{eq:NNclassifier})) for classification; thus it serves as a reference to compare DMP with exDAP.
(3) \textbf{DeViSE+DMP}: We use DeViSE~\cite{DeviseNIPS13} to learn
the visual-semantic embedding into which the power set $P$ is projected.
And we use Eq (\ref{eq:NNclassifier}) for final labelling in the embedding space, i.e., DMP. Thus it corresponds to the extension of \cite{DeviseNIPS13} to multi-label zero-shot learning problems.
(4) \textbf{Mul-DR+exDAP}: Our
Mul-DR is used to learn the visual-semantic embedding, with exDAP
for multi-label classification; thus it can be used to compare Multi-DR with SVR. (5) \textbf{Mul-DR+DMP/TraMP}: Our method with either of the two proposed ZSL algorithms used. For fair comparison, all
results use  self-training strategy in Eq (\ref{eq:self-training}) to update the prototypes.

\begin{figure}
\begin{centering}
\includegraphics[scale=0.35]{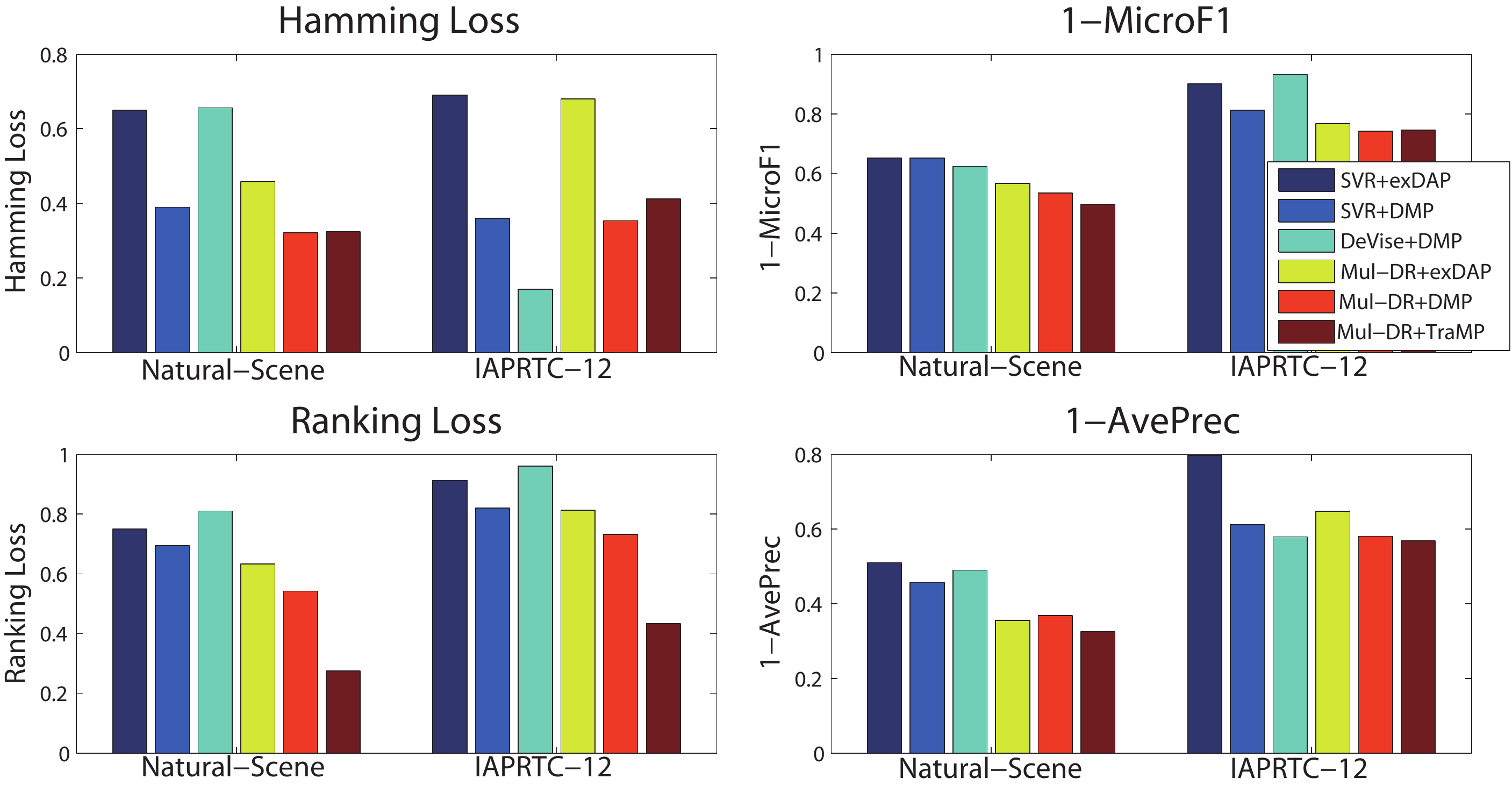}
\par\end{centering}

\protect\caption{\label{fig:Comparing-methods-ofMLdataset}Comparing different zero-shot multi-label classification methods on Natural
Scene and IAPRTC-12.}

\end{figure}

\subsection{Results}

\noindent \textbf{Our Mul-DR model vs.~alternatives}\quad The results obtained by various competitors on Natural-Scene and IAPRTC-12 are shown in Fig.~\ref{fig:Comparing-methods-ofMLdataset}. We first compare our Mul-DR with the alternative SVR and DeViSE model for learning the projection from raw images to the semantic word space. It is evident that our Mul-DR
 significantly improve the results on conventional
SVR \cite{lampert2009zeroshot_dat,lampert13AwAPAMI} regression model (Mul-DR+DMP>SVR+DMP, Mul-DR+exDAP>SVR+exDAP). This is because that SVR treats each of the 100 semantic word space dimensions independently, whilst our multi-output regression model, as well as the DeViSE model \cite{DeviseNIPS13} capture the correlations between different dimensions. Comparing to the DeViSE model \cite{DeviseNIPS13} (Mul-DR+DMP vs. DeViSE+DMP), our regression model is also clearly better using  three of the four evaluation metrics, suggesting that direct and explicit mapping between the image space and the semantic word space is a better strategy. The only case where a better result is obtained by  DeViSE+DMP is on the IAPCTC-12 dataset with Hamming Loss. But this result is worth further discussion. In particular, we note that Hamming Loss treats the false alarm and missing prediction
errors equally. However, for multi-label classification problem, the
distribution of labels is very unbalanced and each image usually has
only a small portion of labels compared to the whole label set. This is particularly the case for IAPCTC-12. The
good result of DeViSE on IAPCTC-12 with better Hamming loss but worse
MicroF1 and Ranking Loss is an indication  that it is mostly predicting no label,
and biased against making any predictions. This explains  the qualitative results of DeViSE shown in Table~\ref{Qualitative_results}.

\noindent \textbf{Our DMP/TraMP vs. exDAP}\quad Given the same regression model, we compared our DAP against the alternative exDAP. The results (SVR+DMP>SVR+exDAP, Mul-DR+DMP>Mul-DR+exDAP) show that our algorithm, which is based on synthesising the label combinations in order to encode the multi-label correlations, is superior to exDAP which treats each label independently and decomposes the multi-label classification problem as multiple single label classification problems. Comparing the two proposed algorithms -- DMP and TraMP, the main difference is that TraMP  transductively exploits the manifold structure of the test data for label prediction.  Figure \ref{fig:Comparing-methods-ofMLdataset} shows that this tranductive label prediction algorithm is better overall. Specifically, TraMP has much better Micro-F1, Ranking Loss and AP than DMP.  The  NN classifier (Eq (\ref{eq:NNclassifier})) used in
DMP is directly minimising the Hamming Loss. This explains why TraMP is slightly worse than DMP on IAPCTC-12 on  Hamming
Loss.

\begin{figure}
\begin{centering}
\includegraphics[scale=0.47]{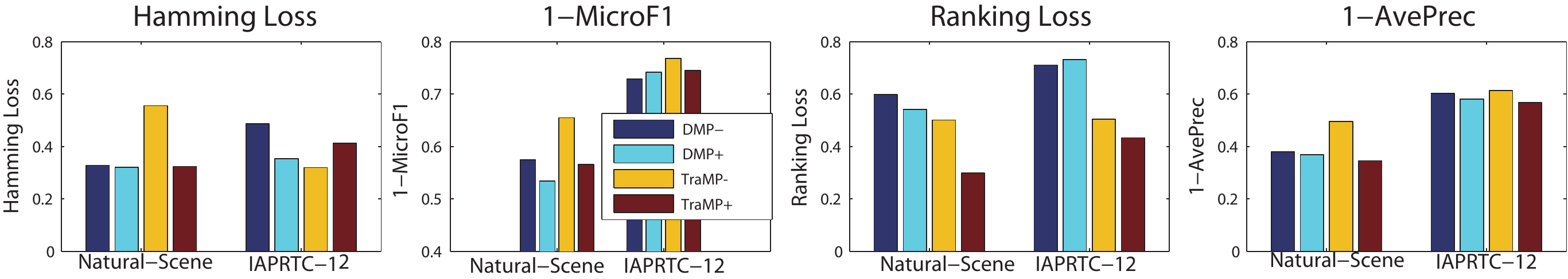}
\par\end{centering}
\protect\caption{\label{fig:Effectiveness-of-self-training}Effectiveness of self-training
on DMP and TraMP.}
\end{figure}

\noindent \textbf{Effectiveness of the self-training step}\quad In this experiment we compare the results of our DMP and TraMP with
and without the self-training step in Eq (\ref{eq:self-training}).
We use `-' and `+' to indicate algorithms without and with self-training
respectively. Both DMP and TraMP use Mul-DR  to infer the
word vector $\hat{Y}_{T}$. As shown in Fig.~\ref{fig:Effectiveness-of-self-training}, the self-training
step clearly has a positive influence on the multi-label prediction performance. This result suggests that this simple step is helpful in making the learned Mul-DR model from the auxiliary data generalise better to the target data.

\noindent \textbf{Qualitative results}\quad Table \ref{Qualitative_results} gives a qualitative comparison of multi-label annotation by
our DMP and TraMP with DeViSE on IAPCTC-12. As discussed, DeViSE is
too conservative on this  dataset and assigns no label to
most instances.

\begin{table}
\begin{centering} \begin{tabular}{c|c|c|c|c} \multicolumn{1}{c|}{} & \includegraphics[width=0.15\textwidth,height=0.2\columnwidth]{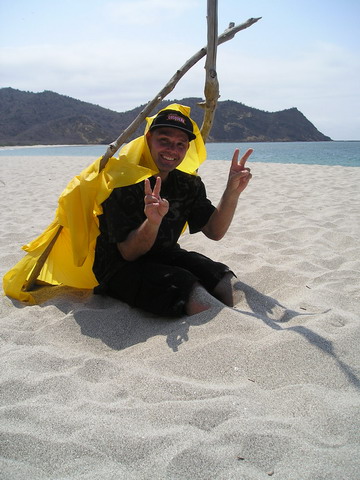} & \includegraphics[width=0.15\textwidth,height=0.2\columnwidth]{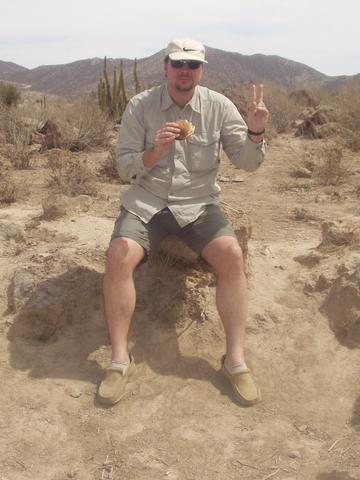} & \includegraphics[width=0.15\textwidth,height=0.2\columnwidth]{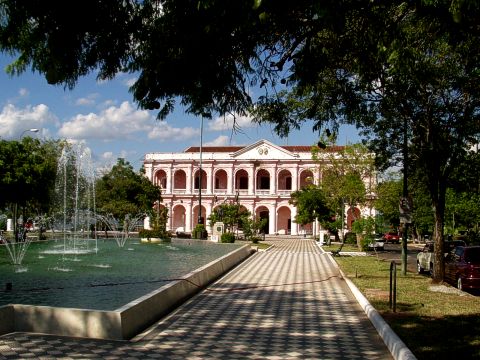} & \includegraphics[width=0.15\textwidth,height=0.2\columnwidth]{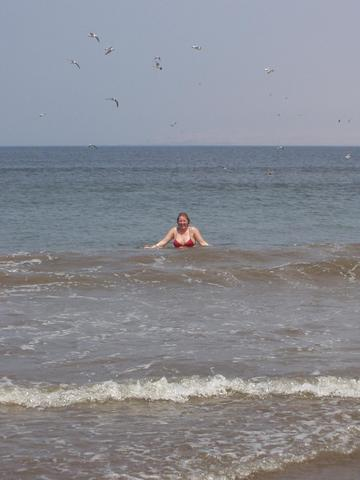}\tabularnewline \hline  \multirow{2}{*}{Groundtruth} & sand-beach,  & landscape-nature, & grass & sand-beach, \tabularnewline  & mountain,sky &  mountain, sky &  & sky\tabularnewline \hline  \multirow{2}{*}{Mul-DR+DMP} & sand-beach,  & landscape-nature,  & grass & sand-beach,\tabularnewline  & sky & mountain, sky &  &  sky\tabularnewline \hline  \multirow{2}{*}{Mul-DR+TraMP} & sand-beach, & landscape-nature,  & grass, ground,  & ground, sky,\tabularnewline  & mountain, sky & mountain, sky & landscape-nature & sand-beach\tabularnewline \hline  \multicolumn{1}{c|}{DeViSE+DMP} & sky & -- & -- & sky\tabularnewline \hline  \end{tabular}
\caption{Examples of  multi-label zero-shot predictions on IAPRTC-12 dataset. Top 8 most frequent labels of landscape-nature branch are considered.}
\label{Qualitative_results}
\par\end{centering}
\end{table}

\vspace{-0.3cm}

\section{Conclusion and future work}

We have  for the first time generalised zero-shot
learning from the single label to the multi-label setting.
It is somewhat surprising that it turns out to be possible to exploit label correlation at test time in the zero shot case -- since there is no dataset of examples to learn co-occurance statistics in the conventional way.
We achieve this via introducing novel strategies to exploit the compositionality of the semantic word
space, and by transductively exploiting the unlabelled test data.

Besides the proposed tailor-made multi-label algorithms -- DMP and TraMP,
our strategy could potentially help other existing multi-label algorithms to generalise to the multi-label zero-shot learning problem. Finally, we note that many prototypes of the power set $P$ actually have an extremely low chance to occur in the test dataset. They  should not be considered in the same way as the other more likely prototypes. Thus another line of ongoing research is to investigate how to  prune low-probability prototypes from the power set $P$.


\bibliography{ref-phd1}

\end{document}